\pdfoutput=1
\documentclass{article}

\usepackage{arxiv_default}
\usepackage{hyperref}
\usepackage{url}

\usepackage[utf8]{inputenc}
\usepackage[T1]{fontenc}
\usepackage{xurl}
\hypersetup{hidelinks}
\usepackage{booktabs}
\usepackage{amsfonts}
\usepackage{amsmath}
\usepackage{amssymb}
\usepackage{microtype}
\usepackage{xcolor}
\usepackage{multirow}
\usepackage{graphicx}
\graphicspath{{figures/}}
\usepackage{subcaption}
\usepackage{placeins}
\usepackage{amsthm}
\usepackage{enumitem}
\usepackage{array}

\theoremstyle{plain}

\newcommand{\ours}{\textsc{Condor}}

\newcommand{\oursfull}{Coupled-Noise Distillation for One-Step Readout}
\newcommand{\oursexpand}{\textbf{Co}upled-\textbf{N}oise \textbf{D}istillation for \textbf{O}ne-Step \textbf{R}eadout}
\newcommand{\mask}{\texttt{[M]}}

\newcommand{\eps}{\varepsilon}

\newcommand{\emask}{e_{\mask}}
\newcommand{\rms}{\mathrm{RMS}}
\newcommand{\CE}{\mathrm{CE}}
\newcommand{\student}{f_\theta}
\newcommand{\teacher}{g_\phi}

\DeclareMathOperator*{\argmax}{arg\,max}
\DeclareMathOperator*{\argmin}{arg\,min}
\newsavebox{\fitwidthbox}
\newcommand{\fitwidth}[1]{%
  \sbox{\fitwidthbox}{#1}%
  \ifdim\wd\fitwidthbox>\linewidth
    \resizebox{\linewidth}{!}{\usebox{\fitwidthbox}}%
  \else
    \usebox{\fitwidthbox}%
  \fi}

\title{A Ticket from Marginals to Joints: Coupled-Noise Distillation for One-Step Block Generation in Diffusion Language Models}

\author{
  Lin Yao$^{1,2}$ \\
  $^1$School of Computer Science, Shanghai Jiao Tong University, Shanghai, 200240, China \\
  $^2$Zhongguancun Academy, Beijing, 100097, China \\
  \texttt{lin.yao@sjtu.edu.cn}
}

\begin{document}

\maketitle
\begin{abstract}
Can a diffusion language model generate a coherent token block in one
forward pass? Masked models already predict every position at once, but
each prediction is the marginal distribution given the visible context,
so the tokens can be mutually inconsistent and later steps revise those
already committed. We introduce \ours{} (\oursfull{}), trained from
scratch to map different noise samples to different coherent blocks.
Initially, random noise is not naturally paired with a target.
Winner-take-all supervision lets different samples specialize, and
self-distillation trains the one-pass output to match the refined
coherent block. TinyStories experiments show diverse, coherent
continuations over successive blocks, one forward pass each. Qualitative
MNIST experiments show that the same approach can extend to multimodal
generation, such as text-to-image and unconditional text-and-image
generation.
\end{abstract}

\section{Introduction}
\label{sec:intro}

\begin{figure}[!ht]
    \centering
    \includegraphics[width=1.0\linewidth]{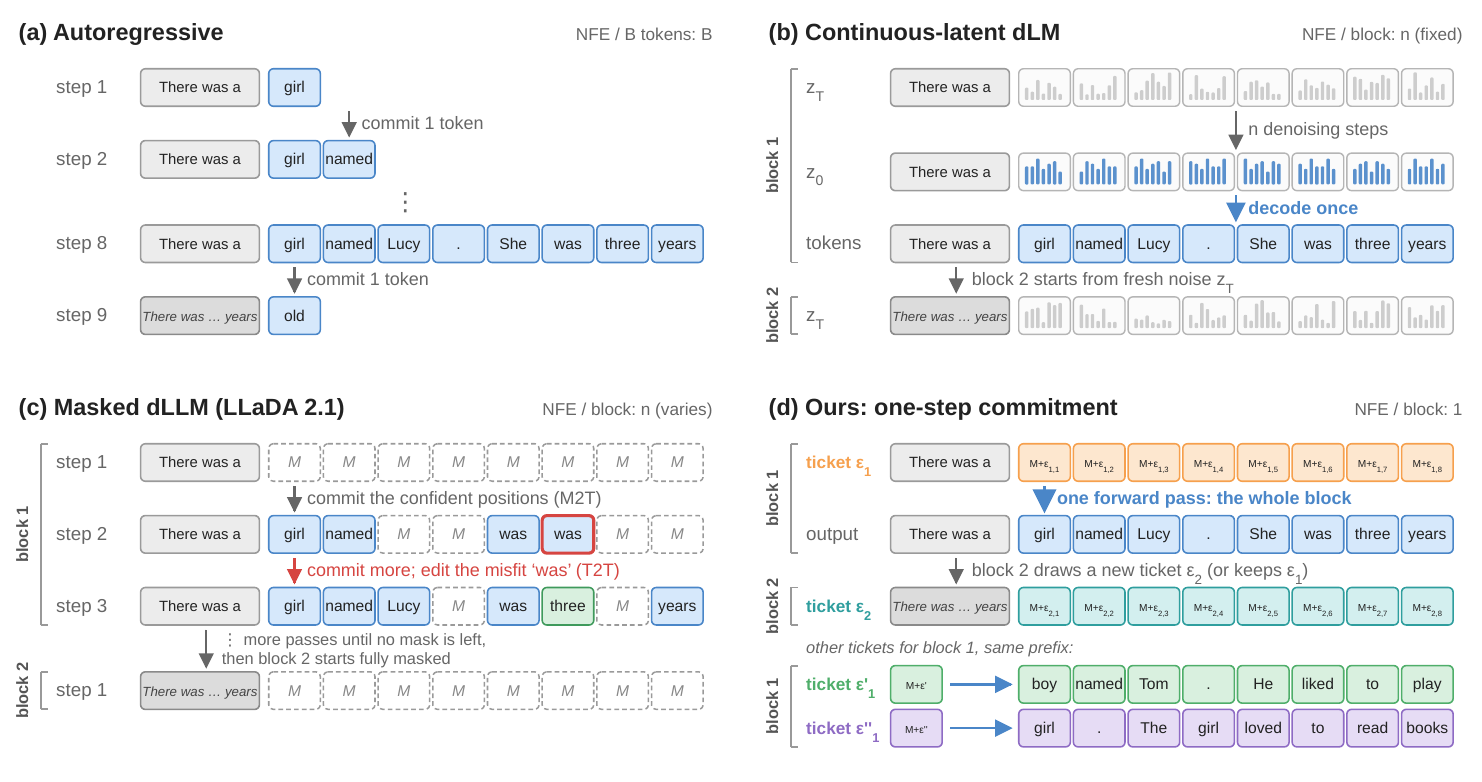}
    \caption{Committing a $B{=}8$-token block.
    \textbf{(a)}~AR, one token per pass.
    \textbf{(b)}~Continuous-latent dLM, $n$ denoising steps.
    \textbf{(c)}~Masked dLLM, commit then edit.
    \textbf{(d)}~Ours, one noise ticket per pass.}
    \label{fig:overview}
\end{figure}

Standard autoregressive (AR) decoding generates one token per forward
pass~\citep{vaswani2017attention,radford2019gpt2}. Conditioning each token
on the preceding tokens supports coherent generation, but makes decoding
sequential: a continuation of $L$ tokens requires $L$ forward passes.
Subsequent work aims to preserve generation quality while improving
decoding efficiency.
Diffusion language models (dLLMs) extend the continuation one block after
another, and improve the generation efficiency by decoding each new block
in fewer than $L$ refinement steps
(Fig.~\ref{fig:overview}). Within a block, two designs are used.
Continuous-latent dLLMs denoise a continuous representation from Gaussian
noise and decode it into tokens at the end.
The continuous latent to be recovered must be obtained before or within
the training procedure: some models take it from a frozen pretrained
model~\citep{hu2026elf}, while others train a text VAE alongside the
diffusion model~\citep{guo2026cola} or learn the embeddings
jointly~\citep{li2022diffusionlm}.
Masked dLLMs, including the LLaDA
series~\citep{nie2025llada,nie2026illada,nie2026llada21}, do not require
this additional continuous latent. They operate directly on discrete
tokens: each step fills several tokens in parallel until no mask remains
in the current block.
Parallel masked predictions add a further difficulty: each is based on the
shared visible context, and tokens filled in the same step cannot see each
other, so the predictions can be mutually
inconsistent~\citep{gu2018nat,kang2025parallelbench}. Later steps have to
remove that inconsistency by revising the visible
tokens~\citep{nie2026llada21,wang2025remdm,remedi2025}.
Both designs still produce one block over several steps.
These limits raise our question: can a dLLM generate a coherent
block in one forward pass?
Especially, can it do so without a continuous latent obtained before or
within training, and without a later revision of the visible tokens?

A masked dLLM already predicts every position of a block in one forward
pass, yet the result need not be a coherent block. Under token-wise
cross-entropy, the optimal prediction at each masked position is its
marginal given the visible tokens~\citep{ou2025radd}, so sampling all
positions at once draws from a product of marginals rather than from the
joint distribution of the block. The obstacle is not the number of steps
but the lack of a per-sample variable that selects one joint mode.
We give a masked dLLM
such a variable by adding a Gaussian noise field $\eps$ to its mask
embeddings. One noise sample (one \emph{ticket}) should select one
coherent block, and different tickets should select different blocks
(Fig.~\ref{fig:overview}d).
Because noise is drawn independently of the data, the model is never
told which block a ticket should map to, so this coupling must be
learned.
We train the dLLM from scratch to learn this coupling.
\ours{} (\oursexpand{}) combines winner-take-all (WTA)
supervision~\citep{guzmanrivera2012mcl,lee2016smcl,rupprecht2017mhp,makansi2019ewta}
with self-distillation
(Fig.~\ref{fig:method}). For a prefix and the
one block about to be generated, we draw $k$ noise samples. These samples
then enter two branches. In the first, a ratio of positions in that block
is masked and the remaining positions keep the ground truth. Each masked
position adds its noise to the mask embedding before the model predicts. Only the prediction closest to the ground
truth in cross-entropy is kept as the loss; the others are discarded.
This symmetry breaking lets different noises learn different
continuations. In the second, the whole block is masked and the $k$
different noises are applied. For each noise ticket, we first iterate
over several steps, filling only part of the mask each time, and obtain
a relatively coherent sequence label. The one-step prediction is trained
to match it, so one forward pass can produce that coherent block. WTA
assigns each noise its own continuation, and self-distillation carries
that continuation into the single pass.

Our contributions are as follows.
\begin{itemize}[nosep,leftmargin=*]
\item We introduce a from-scratch model, \ours{}. By combining WTA
supervision with self-distillation, it becomes the first diffusion
language model that has the capability to generate one complete coherent
token block per forward pass.
\item We achieve coherent block-wise language generation. Each block is committed
in one forward pass, and successive blocks remain a readable
continuation.
\item We achieve the same one-pass commitment in a multimodal setting,
both for text-to-image and for unconditional generation of text and
image, showing that the method generalizes beyond text and has potential
for broader multimodal generation.
\end{itemize}
All experiments are conducted at small scale; our claims concern the
mechanism rather than large-scale performance.

\section{Related Work}
\label{sec:related}

\paragraph{Efficiency and joint coherence trade-off in autoregressive decoding.}
Autoregressive decoding spends one forward pass on each token, so the
number of passes grows with the length of the continuation.
Autoregressive models try to reduce that number of passes in several ways.
Speculative decoding lets a lightweight model draft a few tokens; the
large model checks them in one forward pass and keeps only the prefix it
would itself have continued~\citep{leviathan2023speculative}.
Medusa predicts several tokens from one prefix in a single pass. The
guesses cannot see each other, so the draft can conflict; verification
then keeps the longest prefix consistent with the model's own
predictions~\citep{cai2024medusa}.
Jacobi decoding and Consistency LLMs fill a span of positions at once.
The next round rewrites every position from the previous round's left
context, until each token is the greedy choice given that
context~\citep{santilli2023jacobi,kou2024cllm}.
These methods share a generate-then-self-verify procedure, and they still
require multiple steps.

\paragraph{More efficient token generation with dLLMs.}
Diffusion language models reduce the number of passes in another way.
They extend the continuation one block after another.
A block of $B$ tokens is generated in $t$ iterative steps, with $t<B$.
The two families differ in how that block is generated.
Continuous-latent models denoise exactly as a standard diffusion model
does, from Gaussian noise to a continuous
representation~\citep{li2022diffusionlm,gong2023diffuseq,lovelace2023latent,hu2026elf,guo2026cola,yu2026dsl,zhou2026ccdd}.
That continuous representation is built before or jointly with diffusion
training, and it must be decodable back into tokens.
It is taken from a frozen pretrained model~\citep{hu2026elf}, learned by a
text VAE~\citep{guo2026cola}, or trained jointly as token
embeddings~\citep{li2022diffusionlm}.
Because this denoising follows traditional diffusion, progressive
distillation, consistency models, rectified flow, distribution matching,
and MeanFlow can be expected to reduce the number of denoising steps here
as well, even to one
step~\citep{salimans2022progressive,song2023consistency,liu2023rectified,yin2024dmd,geng2025meanflow}.
To our knowledge, continuous-latent language models have not yet adopted
this further reduction.
Masked models write discrete tokens and fill several positions in a
step~\citep{austin2021d3pm,lou2024sedd,sahoo2024mdlm,shi2024md4,ou2025radd,nie2025llada,ye2025dream,arriola2025bd3lm,nie2026llada21,chang2022maskgit}.
Parallel fills can be mutually inconsistent~\citep{kang2025parallelbench}, so later work
chooses which positions to commit~\citep{wu2025fastdllm}, revises them, or
turns them back into masks on subsequent
passes~\citep{ghazvininejad2019maskpredict,nie2026llada21,wang2025remdm,zhai2026core,remedi2025,kim2025prism,he2025mdpo,schiff2026proseco}.
Some methods introduce distillation from an already trained multi-step
teacher to reduce the number of sampling
steps~\citep{deschenaux2024sdtt,sahoo2025duo,hayakawa2024di4c}.
The sampling schedule becomes shorter, yet still requires multiple
passes, except for DiMO, which has validated one-pass generation on
images~\citep{zhu2025dimo}.
\ours{} is trained from scratch, without a pretrained teacher, and
is capable of committing each block in one forward pass for both text
and images.

\paragraph{A necessary condition for one-step coherence.}
One step can commit several tokens in parallel.
Those tokens form a coherent block only when the input and the output
form a one-to-one or many-to-one pair: one input must select one coherent
combination.
Methods differ in how they form this pair.
(i)~VAE-like methods encode the target $x$ into a high-dimensional latent $z$
and decode $z$ back into $x$, so $z$ and $x$ form the pair inside each
training example~\citep{kingma2014vae}.
An earlier discrete-latent
decoders~\citep{kaiser2018latent},
FlowSeq~\citep{ma2019flowseq}, LaNMT~\citep{shu2020lanmt},
VADD~\citep{xie2026vadd}, and Cola DLM~\citep{guo2026cola} are trained in
this way.
The pair fails when the decoder ignores the latent, as a conditional
generator can ignore an input noise code, or when the posterior collapses
onto the
prior~\citep{isola2017pix2pix,zhu2017bicyclegan,bowman2016vae}.
(ii)~Flow models form the pair from the initial noise.
Rectified flow~\citep{liu2023rectified} and
MeanFlow~\citep{geng2025meanflow} carry one noise sample to one result
in a single forward pass.
ELF~\citep{hu2026elf} is a dLLM trained with flow, so it has the same
one-pass potential, but it does not take that step.
Consistency distillation~\citep{song2023consistency} and
DMD~\citep{yin2024dmd} distill a diffusion model so that one noise sample
yields one result in a single forward pass.
DiMO~\citep{zhu2025dimo} distills a masked diffusion model, taking a
random token initialization as the noise, so that one initialization
yields one result in a single forward pass.
(iii)~Under one condition, IWAE~\citep{burda2016iwae} uses different inputs and obtains different
predictions of the same label. Every prediction remains in the loss.
Instead of token-level cross-entropy, it multiplies the token probabilities
into the sentence probability. This product represents the joint
distribution of the continuation, rather than the per-position marginals.
The weight on each input is a choice: the inputs can be averaged uniformly,
or the weights can be tilted by the loss.
(iv)~Winner-take-all (WTA) losses~\citep{guzmanrivera2012mcl,lee2016smcl,rupprecht2017mhp,makansi2019ewta}
also draw several inputs, but supervise only the
lowest-loss one, so different inputs specialize to different outputs, as
in Social GAN~\citep{gupta2018socialgan} and
LoRA-MCL~\citep{loramcl2025}.
\ours{} uses self-distillation as well as WTA.
Each noise is self-distilled into a different label, so self-distillation
also strengthens learning toward different labels.

\suppressfloats[t]
\section{Method}
\label{sec:method}
\label{sec:noisecond}

We first define $f_\theta$, a block-wise dLLM based on a Transformer. A prefix
$c$ and a continuation block of $B$ tokens are the input of $f_\theta$.
Attention inside a block is bidirectional. Across blocks, a token attends
only to its own block and to earlier blocks~\citep{arriola2025bd3lm}.
During training, each block to be predicted has ground truth $x$, and $x_i$ is its token
at position $i$. $z$ is the current input of that block, formed from $x$
and a mask position set $A$:
\begin{equation}
z_i \;=\; \begin{cases} \mask & i\in A, \\ x_i & i\notin A. \end{cases}
\label{eq:z}
\end{equation}
The masked fraction may be the whole block. Each visible token keeps its
own embedding. The noise can be added at the input, or at an intermediate layer $l$.
The default is the input, $l=0$. The noised state then continues into the next layer.
At each mask position,
\begin{equation}
\tilde e_i \;=\;
\begin{cases}
\emask \;+\; \sigma\cdot\eps_i & l=0, \\
h_i^{(l-1)} \;+\; \sigma\cdot\eps_i & l>0,
\end{cases}
\qquad i\in A ,
\label{eq:noise}
\end{equation}
where $\eps\in\mathbb{R}^{B\times d}$ with $d$ the embedding dimension,
$\emask$ is the mask embedding, $\sigma$ scales the noise and is learned during training,
and $h_i^{(l-1)}$ is the hidden state from the previous layer.
Each row is drawn
$\eps_i\overset{\mathrm{iid}}{\sim}\mathcal N(0,I_d)$, or one vector
$u\sim\mathcal N(0,I_d)$ is copied onto every $i\in A$.
Instead of Eq.~\eqref{eq:noise}, $\tilde e_i$ can be initialized directly as the embedding of a random token.
Table~\ref{tab:main} ablates this initialization.

$\bar{x}_i$ is the vector of logits that $f_\theta$ predicts at position $i$:
\begin{equation}
\bar{x}_i \;=\; f_\theta(c,z,\eps)_i ,
\qquad i\in A .
\label{eq:logits}
\end{equation}

\begin{figure}[!ht]
    \centering
    \includegraphics[width=1.0\linewidth]{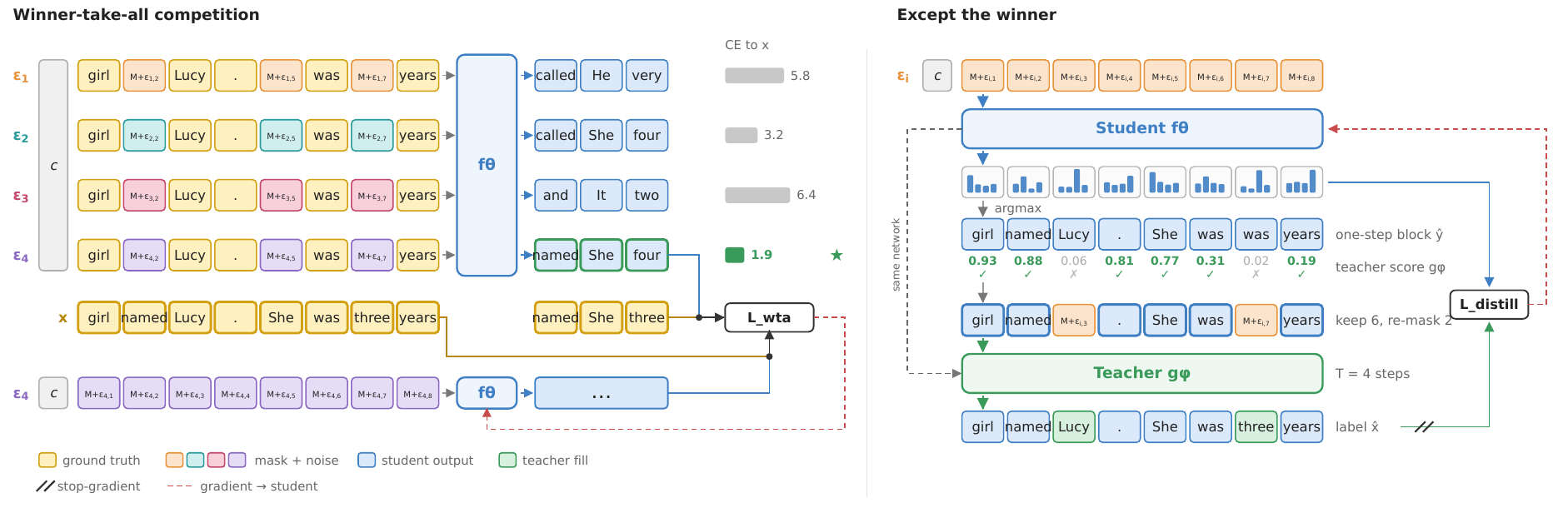}
    \caption{Training \ours{}.
    Left: the winner-take-all task.
    Right: self-distillation.}
    \label{fig:method}
\end{figure}

\ours{} has two training tasks (Fig.~\ref{fig:method}).
(i)~Under a partial mask, $k$ noises each produce a prediction. Every
prediction is scored against the ground truth, but a gradient is sent
only along the lowest-loss noise. The rest are dropped.
The winner noise also initializes a full mask, which is trained against the ground truth.
(ii)~Under a full mask, the other $k-1$ noises are self-distilled to $k-1$ labels,
one label per noise, and the one-pass prediction for a noise is trained
to match its label.

\subsection{Winner-take-all (WTA) on a partial mask}
\label{sec:winner}
On a fixed $(c,z)$ we draw $k>1$ noises
$\eps^{(j)}\in\mathbb{R}^{B\times d}$, $j=1,\dots,k$.
This $z$ is one uniform mask sample, shared by these noises.
The mask count is drawn uniformly from $\{1,\dots,B\}$, and the masked positions are a uniform subset of that size.
For noise $j$, the predicted logits are
$\bar{x}_i^{(j)}=f_\theta(c,z,\eps^{(j)})_i$.
$\mathrm{CE}_j$ is the cross-entropy of these logits against the
ground-truth tokens on the masked positions.
$j_{\mathrm{win}}$ is the index of the noise with the lowest
$\mathrm{CE}_j$.
The partial-mask loss keeps only that cross-entropy:
\begin{equation}
\mathrm{CE}_j
\;=\; \sum_{i\in A}\mathrm{CE}\big(\bar{x}_i^{(j)}, x_i\big) ,
\qquad
j_{\mathrm{win}} \;=\; \argmin_j \mathrm{CE}_j ,
\qquad
\hat{\mathcal{L}}_{\mathrm{wta}}
\;=\; \frac{1}{|A|}\,\mathrm{CE}_{j_{\mathrm{win}}} .
\label{eq:wta}
\end{equation}
Only the winner sends a gradient. The other noises, the \emph{losers},
are left free to stand for other continuations.
Besides this partial mask, the winner noise initializes a full mask, which is also trained against the ground truth.
$z^{\mathrm{full}}$ is that input, Eq.~\eqref{eq:z} with $A=\{1,\dots,B\}$, using the same noise $\eps^{(j_{\mathrm{win}})}$.
$\mathcal{L}_{\mathrm{wta}}$ adds that full-mask cross-entropy to the partial-mask loss:
\begin{equation}
\mathcal{L}_{\mathrm{wta}}
\;=\;
\hat{\mathcal{L}}_{\mathrm{wta}}
+\frac{1}{B}\sum_{i=1}^{B}
\mathrm{CE}\big(f_\theta(c,z^{\mathrm{full}},\eps^{(j_{\mathrm{win}})})_i, x_i\big) .
\label{eq:wta-full}
\end{equation}
$\hat{\mathcal{L}}_{\mathrm{wta}}$ is the partial-mask loss.
$\mathcal{L}_{\mathrm{wta}}$ is that loss together with the ground truth on this full mask.
This winner-selection procedure can also be carried out on a full mask: the same $k$ noises are scored with $A=\{1,\dots,B\}$, and $j_{\mathrm{win}}$ is the lowest full-mask cross-entropy.
Whether the winner comes from the partial mask or the full mask, it is trained against the ground truth.
Table~\ref{tab:main} ablates this choice.

WTA keeps only the lowest-loss noise, so that one noise is trained toward one coherent block.
This is not the only way to supervise the same noises.
Section~\ref{sec:main} tests IWAE, which keeps every noise in the loss, and plain cross-entropy as a comparison that averages them.
It also tests WTA with only the partial-mask loss $\hat{\mathcal{L}}_{\mathrm{wta}}$, without the full-mask term.
Table~\ref{tab:main} reports these results.
Plain cross-entropy averages every noise:
\begin{equation}
\mathcal{L}_{\mathrm{plain}}
=\frac{1}{k|A|}\sum_{j=1}^{k}\mathrm{CE}_j ,
\label{eq:reductions}
\end{equation}
IWAE uses the same mixture, with detached weights tilted by the loss.
Let $c_j=\mathrm{CE}_j/|A|$ be the mean token loss of draw $j$.
\begin{equation}
w_j
= 1 - \frac{c_j - \min_m c_m}{\max_m c_m - \min_m c_m} ,
\qquad
\pi_j = \frac{w_j}{\sum_{m=1}^{k} w_m} ,
\qquad
\mathcal{L}_{\mathrm{iwae}}
= -\frac{1}{|A|}\log\!\left(\sum_{j=1}^{k}\pi_j\, e^{-\mathrm{CE}_j}\right) .
\label{eq:iwae-w}
\end{equation}
No gradient flows into $w_j$ or $\pi_j$.
IWAE with $\pi_j=1/k$ fixes every weight at $1/k$.
Equal draw losses recover this case.
A zero weight removes that draw from the sum.

\subsection{Self-distillation on a full mask}
\label{sec:coupling}

Self-distillation uses two networks, student $\student$ and teacher
$\teacher$.
For a noise $\eps$, the student and the teacher jointly form the
distillation target $\hat{x}$.
The parameters of $\teacher$ can be the same as the current student,
an exponential moving average of the student, or a frozen WTA teacher,
as in Fig.~\ref{fig:method}.
The block is fully masked, so $A=\{1,\dots,B\}$ and every position of
the current input $z$ is a mask. One forward of the student returns
logits.
These logits are greedily discretized into the one-pass tokens.
The probability of token $\bar{y}_i$ on this forward is
\begin{equation}
\bar{x}_i=\student(c,z,\eps)_i,\;
\bar{y}_i=\argmax_v\student(c,z,\eps)_{i,v},\;
p_i=\mathrm{softmax}(\bar{x}_i)_{\bar{y}_i},
\;\; i=1,\dots,B.
\label{eq:onepass-prob}
\end{equation}
The one-pass tokens $\bar{y}$ are not necessarily a good continuation.
The teacher rescores position $i$ on an input
that keeps the other tokens of $\bar{y}$ and replaces position $i$
by the original noised mask $\tilde{e}_i$. This input is written $\bar{y}_{\neg i}$.
$\teacher$ returns logits, and $p_i$ is the softmax probability of the
proposed token:
\begin{equation}
p_i \;=\; \mathrm{softmax}\big(\teacher(c,\bar{y}_{\neg i},\eps)_i\big)_{\bar{y}_i} .
\label{eq:loo}
\end{equation}
Then, by this $p_i$, the $\operatorname{round}(\rho B)$ highest positions
are kept and the rest are remasked.
The teacher is then used to fill the remaining masks.
In each step, the teacher commits the $(1-\rho)B/T$ most confident
argmax tokens until no mask remains, forming the filled block $\hat{x}$.

For each noise $j$, which is the same noise as in the WTA task,
$\bar{x}^{(j)}$ is trained toward $\hat{x}^{(j)}$.
The training loss is
\begin{equation}
\mathcal{L}_{\mathrm{distill}}
=\frac{1}{kB}\sum_{j=1}^{k}\sum_{i=1}^{B}
\mathrm{CE}\big(\bar{x}_i^{(j)},\hat{x}_i^{(j)}\big),
\quad
\mathcal{L}=\mathcal{L}_{\mathrm{wta}}+w(t)\,\mathcal{L}_{\mathrm{distill}},
\label{eq:total}
\end{equation}
with $w(t)$ increased early in training and then kept at $1$.

\subsection{Random-remasking KL self-distillation.}
We also replace the hard-target self-distillation above with a soft-target
variant, while retaining the WTA supervision.
For each loser noise, the student first predicts a proposal from the prefix
and a fully masked block:
\[
\bar x_i^{(j)}=f_\theta(c,z^{\mathrm{full}},\eps^{(j)})_i,
\qquad
\bar y_i^{(j)}=\argmax_v\bar x_{i,v}^{(j)}.
\]
We uniformly select half of the proposal positions to retain and remask
the rest. Writing $K^{(j)}$ for the retained set, the teacher input is
\[
z_i^{(j),\mathrm{keep}}=
\begin{cases}
\bar y_i^{(j)} & i\in K^{(j)},\\
\mask & i\notin K^{(j)}.
\end{cases}
\]
The current student serves as the teacher. One teacher forward uses the
same prefix and noise to produce a target vocabulary distribution at each
remasked position, with no gradient through the teacher:
\[
q_i^{(j)}=\operatorname{stopgrad}\!\left[
\mathrm{softmax}\!\left(g_\phi(c,z^{(j),\mathrm{keep}},\eps^{(j)})_i\right)
\right].
\]
The original full-mask student prediction learns these soft targets using
the per-position loss
\[
\ell_{\mathrm{KL},i}^{(j)}=
D_{\mathrm{KL}}\!\left(q_i^{(j)}\,\|\,
\mathrm{softmax}(\bar x_i^{(j)})\right),
\qquad i\notin K^{(j)}.
\]
The retained tokens condition the teacher and receive no KL loss.
This replaces confidence-ranked retention and iterative hard-token filling
with random retention and one teacher forward.
We evaluate this variant in Table~\ref{tab:main} and the block-wise comparison in Section~\ref{sec:block}.

\paragraph{Hidden-state diversity.}
We also evaluate a repulsion loss on the final hidden states of the
full-mask student forward.
Let $\mathcal C$ be the non-padding content positions in a training sequence,
and let $h_i^{(j)}$ be the output hidden state at position $i$ for noise $j$.
We compare states at the same position across all noise pairs, including
the winner, and minimize the negative mean cosine distance:
\begin{equation}
\mathcal L_{\mathrm{div}}
=-\frac{2}{k(k-1)|\mathcal C|}
\sum_{1\le j<r\le k}\sum_{i\in\mathcal C}
\left(1-
\frac{h_i^{(j)}\cdot h_i^{(r)}}
{\|h_i^{(j)}\|_2\,\|h_i^{(r)}\|_2}\right).
\label{eq:div}
\end{equation}
The loss is averaged over nonempty training sequences, with gradients
through both states in each pair.
When this regularizer is enabled, the total loss is
\begin{equation}
\mathcal L
=\mathcal L_{\mathrm{wta}}
+w(t)\,\mathcal L_{\mathrm{distill}}
+\lambda_{\mathrm{div}}(t)\,\mathcal L_{\mathrm{div}}.
\label{eq:total-div}
\end{equation}
The distillation term uses the remasked-position KL loss for the KL variant
and the hard-target loss otherwise.
The repulsion weight rises linearly from zero to $0.1$ over the first
$5\%$ of training, then falls linearly to zero at the end of the full
training schedule.
We use it in the hidden-state-repulsion ablation and the single-block and block-wise KL runs;
the basic recipe and original block-wise run use zero repulsion weight.

\subsection{Several blocks, and images}
\label{sec:blockmethod}

For block-wise generation inference, the model generates one block in a single
forward pass, appends it to the prefix, and samples fresh noise for the
next block. Each block is conditioned on the given prefix and all
previously generated blocks. Repeating this procedure produces longer
sequences while retaining one forward pass per block.

The same framework also supports multimodal generation, including
text-to-image generation and unconditional generation of text and Images.
In text-to-image generation, the text serves as the prefix
for the image blocks. In unconditional generation, the model first
generates text without a supplied text prefix, then generates the image
conditioned on that text. Images are divided into patches.
When a patch is an input, its pixels pass through a linear layer into the token dimension and form a token.
When a patch is predicted, the model predicts continuous pixel values
within each patch,
and mean squared error replaces cross-entropy in both the winner-take-all
and self-distillation objectives for image prediction.
For image self-distillation, continuous patch predictions do not provide
confidence scores for ranking patches. We therefore randomly select
which patches of the one-pass prediction to remask, and randomly select
which remaining masked positions to fill at each teacher step.

\section{Experiments}
\label{sec:exp}

\subsection{Experimental setup}
\label{sec:setup}

We evaluate \ours{} on four tasks spanning text and image generation,
using a single forward pass per block throughout.
\emph{Single-block text continuation} generates an eight-token continuation conditioned on an
eight-token prefix.
\emph{Multi-block text continuation} extends a text prompt with eight consecutive blocks of
eight tokens each.
\emph{Word-to-image} generates a digit image conditioned on its English
class name (\emph{zero}--\emph{nine}).
\emph{Unconditional} generates a digit word followed by its corresponding
image.

\paragraph{Data.}
Both text tasks use TinyStoriesV2-GPT4~\citep{eldan2023tinystories}
and the same byte-level BPE of vocabulary $8{,}192$ fit on the TinyStories training set.
\emph{Single-block text continuation} is sampled at sentence level.
Each training window has $16$ tokens, which is $2$ blocks of eight.
A sentence shorter than this window is filled with the following sentences.
\emph{Multi-block text continuation} is sampled at document level,
packed into sequences of up to $1024$ tokens, which is $128$ blocks of eight.
A document shorter than this length is filled with the following documents.
Both image tasks use MNIST, with each $28\times28$ image partitioned into
sixteen $7\times7$ patches.

\paragraph{Models and training.}
We train each task from scratch with AdamW, using the setup in
Table~\ref{tab:training-config}.
The \ours{} choices are the same for all four tasks.
Word-to-image and unconditional share one setting.
Each update draws $k=4$ noises and selects the winner by cross-entropy against the
ground truth on the partial mask.
That winner is trained against the ground truth on the partial mask and on the full mask; the other
$k-1$ noises are trained on self-distillation targets.
The teacher is the current student, with $\rho=0.75$ and $T=2$.
One noise vector is shared inside each block and added at $l=0$.
$\sigma$ is learned, initialized at $0.01$, and floored at $0.005$.
Additional optimizer details are in Appendix~\ref{app:training-details}.

\begin{table}[!ht]
\centering\small
\caption{Experimental settings.
Word-to-image and unconditional share the image column.
A merged entry applies to all four tasks.
The single-block column is the Base row of Table~\ref{tab:main}.}
\label{tab:training-config}
\setlength{\tabcolsep}{3.5pt}
\fitwidth{\begin{tabular}{@{}lccc@{}}
\toprule
& \shortstack{Single-block\\text}
& \shortstack{Multi-block\\text}
& \shortstack{Word-to-image\\Unconditional} \\
\midrule
Architecture & \multicolumn{3}{c}{Block-causal Transformer} \\
Learning rate & \multicolumn{3}{c}{$3\times10^{-4}$} \\
Global batch size & \multicolumn{3}{c}{$512$} \\
Tokenizer & \shortstack{Byte-level BPE\\$8{,}192$} & \shortstack{Byte-level BPE\\$8{,}192$} & LLaDA \\
Parameters & $8.39$M & $291.6$M & $5.89$M \\
Layers / hidden / FFN & $6/256/1024$ & $24/1024/2816$ & $6/256/512$ \\
Attention heads (query / key-value) & $8/8$ & $8/4$ & $8/2$ \\
Sequence / block length & $16/8$ & $1024/8$ & $16/8$ \\
Total updates & $100{,}000$ & $10{,}000$ & $100{,}000$ \\
\bottomrule
\end{tabular}}
\end{table}

\paragraph{Evaluation.}
\label{sec:metrics}
For fair comparisons, directly compared methods use matched evaluation
examples and scoring protocols within their respective task settings.

For \emph{Single-block text continuation}, we give $N$ prefixes and continue from each one.
The prefixes are $b_{i,j}$, block $j$ of sentence $i$, with $i=0,\ldots,N-1$ and $j=0$, the first block.
For each $b_{i,0}$ we generate $m$ next blocks $b_{i,1,m}$, where the last index marks the generation.
Block-wise text continuation has two settings.
For \emph{Free continuation}, we give $N$ prefixes $b_{i,0}$, $i=0,\ldots,N-1$, and continue from each one.
For each $b_{i,0}$ we draw $m$ continuations, and continuation $m$ writes the next $w$ blocks $b_{i,1,m},\ldots,b_{i,w,m}$.
$b_{i,k,m}$ is written from $b_{i,0}$ together with $b_{i,1,m},\ldots,b_{i,k-1,m}$ of that continuation.
For \emph{GT prefix}, we give each sentence $i=0,\ldots,N-1$ and each length $k=1,\ldots,w$ the prefix $b_{i,0},\ldots,b_{i,k-1}$ from the original sentence.
All $m$ noises see this same prefix, and noise $m$ generates the current block $b_{i,k,m}$.

A block is valid when GPT-6-Astra judges it grammatically and contextually acceptable, unless stated otherwise.
Natural truncation is allowed, and an earlier error in a free continuation does not by itself invalidate the current block.
$v_i$ is the number of valid blocks among the $m$ blocks of sentence $i$.
$u_i$ is the number of distinct blocks among these $m$ blocks of sentence $i$.
$(u\&v)_i$ is the number of distinct and valid blocks among the $m$ blocks of sentence $i$.
Two blocks are the same when every token matches.
A repeated block adds $1$ to $u_i$ and every copy to $m$; the same split applies to $(u\&v)_i$ and $v_i$.
\begin{gather*}
V=\frac1N\sum_i\frac{v_i}{m},\quad
V_{\ge 1}=\frac1N\sum_i\mathbf{1}[v_i\ge 1],\quad
U=\frac1N\sum_i\frac{u_i}{m},\\
U\&V=\frac1N\sum_i\frac{(u\&v)_i}{m},\quad
\frac{U\&V}{V}=\frac{\sum_i(u\&v)_i}{\sum_i v_i}.
\end{gather*}
Validity $V$ is the fraction of valid blocks, and Uniqueness $U$ is the fraction of distinct blocks.
$V_{\ge 1}$ is the fraction of sentences with $v_i\ge 1$.
The five rates are percentages.

Besides these five rates, \emph{gen-PPL} is the GPT-2-large perplexity of the generated tokens given the conditioning prefix.
The judging rubric appears in Appendix~\ref{app:judging}.

The single-block table uses $N=256$ and $m=16$, with prefix $b_{i,0}$.
The block-wise table uses $N=4$, $m=4$, and $w=8$.
Validity, $V_{\ge 1}$, Uniqueness \& Validity, and their ratio use the first $32$ sentences ($512$ blocks).
Uniqueness and gen-PPL use all $256$ sentences, so that Uniqueness column is not the numerator of the ratio.

\subsection{Single-block text continuation}
\label{sec:exp8}
\label{sec:main}

Table~\ref{tab:main} is a fair comparison on eight-token prefixes generating the next eight tokens.
Every row is noise-initialized, except the random-token rows.

The ground-truth-only rows are trained without self-distillation.
Aside from \emph{WTA on the partial and the full mask}, training places $k$ noises on a partial mask, and the rows differ only in the loss.
That \emph{WTA on the partial and the full mask} row also trains the winner against the ground truth on the full mask.
These rows test whether training works without distillation; under one-pass decoding, their metrics are lower than the rows that include it.
Confidence-order WTA reuses the $k=4$ checkpoint and commits the most confident remaining token on each of eight forwards, holding its noise fixed; all other rows use one forward.

The one-step adaptations are other methods. Details are in Appendices~\ref{app:dimo} and~\ref{app:flow}.
DiMO is a one-step distillation method from images, which we transfer to text.
Its Uniqueness is high, so the generated blocks are almost all different, but its validity and U\&V are lower than ours.
MeanFlow and rectified flow are lower still.

The Gaussian rows ablate components of our method.
The mask-embedding-zero row sets the mask embedding to zero.
The independent-noise row draws an independent noise at each position.
The $l=3$ row adds the noise at an intermediate layer.
The hidden-state-repulsion row forces the output hidden states of the $k$ noises apart from one another. Validity falls and Uniqueness rises.
The $k=2$ and $k=8$ rows change the number of candidates.
The batch rows change the batch size.
The select-winner-on-full-mask row selects the winner on the full mask.

The EMA row sets the teacher to an EMA of the student.
The frozen-WTA-teacher row sets the teacher to a frozen WTA model.
None of these choices collapses.
The random-remasking KL row uses the soft-target variant with random retention $\rho=0.5$ and peak repulsion weight $0.1$.
The $\rho=0.5$, $T=2$ row has the student fill half the block and the teacher take two steps.
The $\rho=0.5$, $T=4$ row has the student fill half the block and the teacher take four steps.
The $\rho=0$, $T=1$ row has the teacher write one token at a time, by confidence.
The $\rho=0$, $T=1$, left-to-right row has the teacher write one token at a time, from left to right.

The random-token Base row initializes the block with one repeated token instead of noise.
The independent-token row draws an independent token at each position.
The whole-block-argmax row changes the self-distillation target used in training: it replaces every position by its argmax token and applies this replacement to the whole block three times.
Validity is high, but the blocks are highly repetitive.

\begin{table}[!ht]
\centering\small
\caption{Validity for DiMO, MeanFlow, and rectified flow is the Codex review in Appendices~\ref{app:dimo} and~\ref{app:flow}; confidence-order WTA and hidden-state repulsion use the same locked review. Random-remasking KL uses a separate Codex review with prior exposure to its configuration; labels were locked before aggregation.}
\label{tab:main}
\setlength{\tabcolsep}{1.5pt}
\fitwidth{\begin{tabular}{@{}lrrrrrrr@{}}
\toprule
Method / variant & NFE & \shortstack{Validity\\(\%) $\uparrow$} & \shortstack{$V_{\ge 1}$\\(\%) $\uparrow$} & \shortstack{Uniqueness\\(\%) $\uparrow$} & \shortstack{U\&V\\(\%) $\uparrow$} & \shortstack{(U\&V)/V\\(\%) $\uparrow$} & gen-PPL $\downarrow$ \\
\midrule
\multicolumn{8}{l}{\emph{Ground truth only}} \\
Plain CE & 1 & 5.66 & 6.25 & 11.18 & 0.39 & 6.90 & 325.17 \\
IWAE & 1 & 7.81 & 56.25 & 83.40 & 6.05 & 77.50 & 235.03 \\
IWAE with $\pi_j=1/k$ & 1 & 7.23 & 46.88 & 76.05 & 5.08 & 70.27 & 237.63 \\
WTA, $k=4$ & 1 & 6.25 & 46.88 & 85.99 & 5.08 & 81.25 & 233.87 \\
WTA, $k=4$, confidence order & 8 & 97.66 & 100.00 & 87.70 & 84.38 & 86.40 & 19.63 \\
WTA, $k=8$ & 1 & 10.94 & 65.62 & 93.55 & 9.18 & 83.93 & 189.92 \\
WTA, partial and full mask & 1 & 8.79 & 37.50 & 68.09 & 4.49 & 51.11 & 262.99 \\
\midrule
\multicolumn{8}{l}{\emph{One-step adaptations}} \\
DiMO, hybrid & 1 & 58.40 & 100.00 & 99.51 & 58.20 & 99.67 & 58.65 \\
MeanFlow & 1 & 1.37 & 9.38 & 42.92 & 0.59 & 42.86 & 275.97 \\
Rectified flow & 1 & 1.76 & 18.75 & 42.55 & 1.37 & 77.78 & 248.48 \\
\midrule
\multicolumn{8}{l}{\emph{Gaussian initialization}} \\
Base & 1 & 81.84 & 100.00 & 69.75 & 48.05 & 58.71 & 28.10 \\
Hidden-state repulsion & 1 & 78.71 & 100.00 & 72.09 & 51.95 & 66.00 & 28.24 \\
Mask embedding zero & 1 & 80.86 & 100.00 & 65.70 & 47.07 & 58.21 & 29.28 \\
Independent noise per position & 1 & 80.66 & 100.00 & 80.64 & 63.28 & 78.45 & 33.69 \\
Select winner on full mask & 1 & 70.90 & 100.00 & 42.65 & 20.51 & 28.93 & 29.49 \\
$l=3$ & 1 & 81.84 & 100.00 & 56.93 & 41.02 & 50.12 & 26.46 \\
$k=2$ & 1 & 62.50 & 93.75 & 44.02 & 21.68 & 34.69 & 40.39 \\
$k=8$ & 1 & 86.72 & 100.00 & 82.96 & 67.58 & 77.93 & 26.84 \\
Batch $1{,}024$, $50{,}000$ updates & 1 & 82.42 & 100.00 & 69.07 & 51.56 & 62.56 & 26.45 \\
Batch $256$, $200{,}000$ updates & 1 & 84.57 & 100.00 & 72.39 & 56.64 & 66.97 & 33.78 \\
\midrule
\multicolumn{8}{l}{\emph{Teacher}} \\
EMA $0.9999$ & 1 & 83.59 & 100.00 & 70.39 & 53.71 & 64.25 & 27.97 \\
Frozen WTA teacher & 1 & 84.38 & 100.00 & 63.70 & 46.29 & 54.86 & 25.00 \\
\midrule
\multicolumn{8}{l}{\emph{Self-distillation}} \\
Random-remasking KL & 1 & 73.24 & 100.00 & 57.03 & 32.62 & 44.53 & 22.72 \\
$\rho=0.5$, $T=2$ & 1 & 78.91 & 100.00 & 74.80 & 50.98 & 64.60 & 26.78 \\
$\rho=0.5$, $T=4$ & 1 & 77.73 & 100.00 & 78.64 & 55.27 & 71.11 & 25.37 \\
$\rho=0$, $T=1$ & 1 & 36.33 & 93.75 & 80.64 & 25.59 & 70.43 & 70.03 \\
$\rho=0$, $T=1$, left to right & 1 & 34.38 & 96.88 & 87.21 & 26.76 & 77.84 & 73.36 \\
\midrule
\multicolumn{8}{l}{\emph{Random-token initialization}} \\
Base & 1 & 86.52 & 100.00 & 68.58 & 51.56 & 59.59 & 25.04 \\
Independent token per position & 1 & 71.88 & 100.00 & 77.29 & 50.39 & 70.11 & 35.26 \\
Whole-block argmax, three times & 1 & 85.55 & 96.88 & 22.00 & 16.60 & 19.41 & 24.26 \\
\bottomrule
\end{tabular}}
\end{table}

\subsection{Block-wise text generation}
\label{sec:block}

Table~\ref{tab:block} reports the two self-distillation variants on the
same four TinyStories inputs, with four draws per input and eight
eight-token blocks under Free continuation and GT prefix.
Both checkpoints are evaluated at $10{,}000$ updates. KL labels come
from a separate Codex review, so this comparison spans different
self-distillation recipes and review rounds.

\begin{table}[!ht]
\centering\small
\caption{Block-wise generation with original self-distillation
($10{,}000$ updates) and random-remasking KL ($10{,}000$ updates),
using the $8{,}192$-token BPE and sequence length $1024$.
Each position has $4\times4$ draws; all five rates are percentages.
Missing blocks after EOS count as invalid and as an empty token-ID sequence
in Uniqueness. The ratio is undefined with no valid block.
Gen-PPL pools scored continuation tokens across all eight positions.}
\label{tab:block}
\setlength{\tabcolsep}{2pt}
\fitwidth{\begin{tabular}{lrrrrrrrrrr}
\toprule
& \multicolumn{5}{c}{Free continuation} & \multicolumn{5}{c}{GT prefix} \\
\cmidrule(lr){2-6}\cmidrule(lr){7-11}
Block & Validity & $V_{\ge 1}$ & Uniqueness & U\&V & (U\&V)/V & Validity & $V_{\ge 1}$ & Uniqueness & U\&V & (U\&V)/V \\
\midrule
\multicolumn{11}{l}{\emph{Original self-distillation, step $10{,}000$}} \\
1 & 100.00 & 100.00 & 62.50 & 62.50 & 62.50 & 100.00 & 100.00 & 62.50 & 62.50 & 62.50 \\
2 & 100.00 & 100.00 & 81.25 & 81.25 & 81.25 & 87.50 & 100.00 & 87.50 & 75.00 & 85.71 \\
3 & 87.50 & 100.00 & 93.75 & 81.25 & 92.86 & 81.25 & 100.00 & 68.75 & 50.00 & 61.54 \\
4 & 75.00 & 100.00 & 100.00 & 75.00 & 100.00 & 93.75 & 100.00 & 81.25 & 75.00 & 80.00 \\
5 & 87.50 & 100.00 & 100.00 & 87.50 & 100.00 & 62.50 & 100.00 & 93.75 & 56.25 & 90.00 \\
6 & 75.00 & 100.00 & 100.00 & 75.00 & 100.00 & 68.75 & 100.00 & 81.25 & 56.25 & 81.82 \\
7 & 87.50 & 100.00 & 100.00 & 87.50 & 100.00 & 87.50 & 100.00 & 87.50 & 75.00 & 85.71 \\
8 & 62.50 & 100.00 & 100.00 & 62.50 & 100.00 & 100.00 & 100.00 & 62.50 & 62.50 & 62.50 \\
gen-PPL & \multicolumn{5}{c}{9.98} & \multicolumn{5}{c}{10.80} \\
\midrule
\multicolumn{11}{l}{\emph{Random-remasking KL, step $10{,}000$}} \\
1 & 93.75 & 100.00 & 75.00 & 68.75 & 73.33 & 93.75 & 100.00 & 75.00 & 68.75 & 73.33 \\
2 & 100.00 & 100.00 & 100.00 & 100.00 & 100.00 & 100.00 & 100.00 & 81.25 & 81.25 & 81.25 \\
3 & 93.75 & 100.00 & 100.00 & 93.75 & 100.00 & 87.50 & 100.00 & 68.75 & 56.25 & 64.29 \\
4 & 68.75 & 100.00 & 100.00 & 68.75 & 100.00 & 68.75 & 100.00 & 87.50 & 56.25 & 81.82 \\
5 & 68.75 & 100.00 & 100.00 & 68.75 & 100.00 & 75.00 & 100.00 & 87.50 & 62.50 & 83.33 \\
6 & 62.50 & 100.00 & 100.00 & 62.50 & 100.00 & 75.00 & 100.00 & 68.75 & 43.75 & 58.33 \\
7 & 93.75 & 100.00 & 100.00 & 93.75 & 100.00 & 81.25 & 100.00 & 87.50 & 68.75 & 84.62 \\
8 & 93.75 & 100.00 & 100.00 & 93.75 & 100.00 & 93.75 & 100.00 & 62.50 & 56.25 & 60.00 \\
gen-PPL & \multicolumn{5}{c}{9.06} & \multicolumn{5}{c}{10.03} \\
\bottomrule
\end{tabular}}
\end{table}

Both self-distillation variants can generate longer sentences and stories
block by block. Table~\ref{tab:block} reports the quantitative results,
and Table~\ref{tab:cases} shows complete generated continuations.

\paragraph{Parallel candidate selection.}
For long text, a small failure probability at each block compounds over
successive blocks. Our aim is to use parallel candidates to raise the
validity of each committed block while preserving one decoding round per
block. Let $q_{n,t}$ be the probability that the selected block at step $t$
is valid, conditional on all preceding blocks being valid. The probability
that the first $M$ blocks all pass is
\[
R_n(M)=\prod_{t=1}^{M}q_{n,t},
\qquad
R_n(M)=q_n^M\quad\text{if }q_{n,t}=q_n\text{ at every step}.
\]

At each block, $n$ independent noises initialize $n$ fully masked inputs
with the same prefix, and $f_\theta$ evaluates them in one batched forward.
Each candidate takes the argmax token at every position.
Writing $\bar\ell$ and $\ell_{\min}$ for the mean and minimum log-probabilities
of its eight predicted tokens, we select the whole candidate maximizing
$\bar\ell$, $\ell_{\min}$, or $(\bar\ell+\ell_{\min})/2$;
ties keep the first candidate.
The selected block is appended and the next block draws fresh noises.
Increasing $n$ spends memory and aggregate computation within the same
decoding round. Wall-clock latency depends on the available parallel
hardware and is not benchmarked here.

If independent candidates each have validity $p$ at every valid prefix,
and a perfect selector chooses a valid candidate whenever one exists,
then the ideal per-block and cumulative success rates are
\[
q_n^{\star}=1-(1-p)^n,
\qquad
R_n^{\star}(M)=\bigl[1-(1-p)^n\bigr]^M.
\]
For example, with single-candidate validity $90\%$ and eight candidates,
the ideal probability that all $125{,}000$ eight-token blocks pass,
covering one million generated tokens, is approximately $99.875\%$.
This motivates parallel candidate selection for reliable long text;
practical confidence scores must also select accurately, and validity can
change with the generated prefix.

We therefore vary $n\in\{1,4,8,16\}$ and the three scores while holding
the checkpoint fixed. Table~\ref{tab:block-selection} reports both
per-block validity and measured cumulative success for $M=1,\ldots,8$,
computed directly from each trajectory's frozen labels.
For the original checkpoint, per-block validity rises from $84.38\%$
at $n=1$ to $100\%$ with 16 candidates and either the minimum or
mean--minimum score; the first-eight-block success rate rises from
$37.50\%$ to $100\%$.
The KL sweep is reported in Appendix~\ref{app:block-selection}.
These 16-trajectory results test the effect of parallel candidate count
and selection strategy; they do not establish million-token reliability.

\begin{table}[!ht]
\centering\small
\caption{Effect of parallel candidates and selection strategy with a fixed original checkpoint (step $10{,}000$). Each condition has 16 Free trajectories. Block validity pools the 128 blocks at positions 1--8; the remaining columns give the fraction of trajectories whose first $M$ blocks all pass. Each block contains eight generated tokens, and all rates are percentages.}
\label{tab:block-selection}
\setlength{\tabcolsep}{4pt}
\fitwidth{\begin{tabular}{rlrrrrrrrrr}
\toprule
& & & \multicolumn{8}{c}{First $M$ blocks all valid (\%)} \\
\cmidrule(lr){4-11}
$n$ & Score & \shortstack{Block validity\\(\%)} & $M=1$ & $2$ & $3$ & $4$ & $5$ & $6$ & $7$ & $8$ \\
\midrule
1 & Any & 84.38 & 100.00 & 100.00 & 87.50 & 68.75 & 62.50 & 50.00 & 50.00 & 37.50 \\
4 & $\bar\ell$ & 96.88 & 100.00 & 100.00 & 100.00 & 100.00 & 100.00 & 87.50 & 81.25 & 81.25 \\
4 & $\ell_{\min}$ & 96.09 & 100.00 & 100.00 & 100.00 & 93.75 & 93.75 & 93.75 & 87.50 & 87.50 \\
4 & $(\bar\ell+\ell_{\min})/2$ & 98.44 & 100.00 & 100.00 & 100.00 & 93.75 & 93.75 & 93.75 & 87.50 & 87.50 \\
8 & $\bar\ell$ & 98.44 & 100.00 & 100.00 & 100.00 & 100.00 & 93.75 & 87.50 & 87.50 & 87.50 \\
8 & $\ell_{\min}$ & 97.66 & 100.00 & 100.00 & 100.00 & 100.00 & 100.00 & 93.75 & 93.75 & 81.25 \\
8 & $(\bar\ell+\ell_{\min})/2$ & 98.44 & 100.00 & 100.00 & 100.00 & 100.00 & 100.00 & 93.75 & 93.75 & 87.50 \\
16 & $\bar\ell$ & 98.44 & 100.00 & 100.00 & 100.00 & 93.75 & 93.75 & 93.75 & 87.50 & 87.50 \\
16 & $\ell_{\min}$ & 100.00 & 100.00 & 100.00 & 100.00 & 100.00 & 100.00 & 100.00 & 100.00 & 100.00 \\
16 & $(\bar\ell+\ell_{\min})/2$ & 100.00 & 100.00 & 100.00 & 100.00 & 100.00 & 100.00 & 100.00 & 100.00 & 100.00 \\
\bottomrule
\end{tabular}}
\end{table}

The first three complete stories in Table~\ref{tab:cases} are Free
continuations to EOS from the partial-mask checkpoint at step $10{,}000$.
Each new eight-token block is conditioned on the previously generated
blocks. We draw eight or sixteen noise candidates per block and select
the candidate with the highest model log-probability score.
Here $\bar\ell$ and $\ell_{\min}$ denote the mean and minimum log-probability
of the eight predicted tokens in a candidate block, respectively.
All candidates for a block are evaluated in one batched forward pass.

The last story is a selected complete KL continuation with 8 candidates
and the mean score, showing Lucy asking permission to climb a tree and
thanking her mother.

\begin{table}[!ht]
\centering
\small
\caption{Qualitative text-generation examples.
Top: one-pass continuations of the same eight-token prefix; underlining
marks errors. Below: four selected complete rollouts; the first three use
the original partial-mask checkpoint at step $10{,}000$, and the last uses
random-remasking KL at step $10{,}000$. Headers give per-block noise candidate
counts and selection scores. Bold text marks the initial eight-token input;
the remaining text is generated through EOS. Each candidate uses one forward pass.}
\label{tab:cases}
\begin{tabular}{@{}p{0.48\linewidth}p{0.48\linewidth}@{}}
\toprule
WTA & \ours{} \\
\midrule
\multicolumn{2}{@{}l@{}}{\textbf{Prefix:} \textit{He ran all the way home and gave}} \\
\textit{\underline{the a.. He mom, and}} \newline
\textit{\underline{the a big.....}}
& \textit{his mom big hug. He was very} \newline
\textit{the big, red ball.\newline The} \\
\midrule
\multicolumn{2}{@{}p{0.98\linewidth}@{}}{\textbf{8 noise candidates; score $(\bar\ell+\ell_{\min})/2$.} \newline
\textit{\textbf{Once upon a time, in a warm} little house, there lived a girl named Mia. Mia had a big, red ball that she loved to play with. One day, Mia's mom said, ``Mia, let's go to the park!'' At the park, Mia played with her ball. She threw it up in the air and caught it. But then, she saw a big, friendly dog. The dog said, ``Woof, can I play with your ball too?'' Mia said, ``Yes, let's play together!'' They played with the ball and had lots of fun. When it was time to go home, Mia said, ``Thank you, dog, for playing with me!'' The dog wagged its tail and barked, ``You're welcome, Mia!'' Mia and the dog walked back to the warm little house, happy to play together.}} \\
\midrule
\multicolumn{2}{@{}p{0.98\linewidth}@{}}{\textbf{16 noise candidates; score $\ell_{\min}$.} \newline
\textit{\textbf{One morning, a cat named Tom woke} up early. He thought, ``I want to go to the park.'' He asked his mom, ``Mom, can we go to the park?'' His mom said, ``Yes, let's go!'' At the park, they saw a big tree. The tree had a hole in it. Tom and his mom looked inside the hole. They found a little bird. The bird said, ``Hello, I am a magic bird!'' Tom and his mom were very surprised. The magic bird said, ``I will give you a wish!'' Tom thought for a moment and said, ``I wish for a big ice cream!'' The magic bird made a big ice cream appear. Tom and his mom ate the ice cream and had a great day at the park.}} \\
\midrule
\multicolumn{2}{@{}p{0.98\linewidth}@{}}{\textbf{16 noise candidates; score $\bar\ell$.} \newline
\textit{\textbf{One morning, a cat named Tom woke} up early. He thought, ``I want to find a big cake!'' So, he went to the kitchen. He looked and looked, but he did not find a cake. Then, Tom saw a cake on the table. He thought, ``I want to eat this cake!'' He started to eat the cake. But then, something unexpected happened. The cake started to talk! It said, ``Please don't eat me! I am a magic cake!'' Tom was very surprised. He did not want to eat a magic cake. So, he said, ``Okay, I will not eat you.'' They became friends and played together. From that day on, Tom and the magic cake had lots of fun.}} \\
\midrule
\multicolumn{2}{@{}p{0.98\linewidth}@{}}{\textbf{KL, step $10{,}000$; 8 noise candidates; score $\bar\ell$ (prompt 002, draw 3).} \newline
\textit{\textbf{Once upon a time there was a little} girl called Lucy. She was three years old. One day, Lucy went to the park to play. She saw a big tree and wanted to climb it. She asked her mom, ``Can I climb the tree?'' Her mom smiled and said, ``Yes, but be careful!'' So Lucy started to climb the tree. She went higher and higher until she reached the top. Lucy said, ``This is fun!'' Her mom laughed and said, ``Yes, it is!'' Lucy was very happy. She said, ``Thank you, Mom!'' Her mom smiled and said, ``You're welcome, Lucy!'' The end.}} \\
\bottomrule
\end{tabular}
\end{table}

\subsection{Multimodal generation}
\label{sec:mnist}

\begin{figure}[!ht]
    \centering
    \begin{subfigure}[t]{0.49\linewidth}
        \centering
        \vspace{0pt}
        \includegraphics[width=\linewidth]{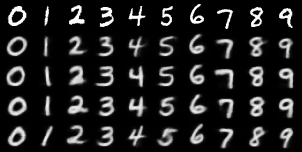}
        \caption{Text-to-image generation (step $18{,}000$). Columns are
                \emph{zero}--\emph{nine}; each sample row uses the same noise across words.}
        \label{fig:mnist}
    \end{subfigure}\hfill
    \begin{subfigure}[t]{0.49\linewidth}
        \centering
        \vspace{0pt}
        \includegraphics[height=0.503311258\linewidth]{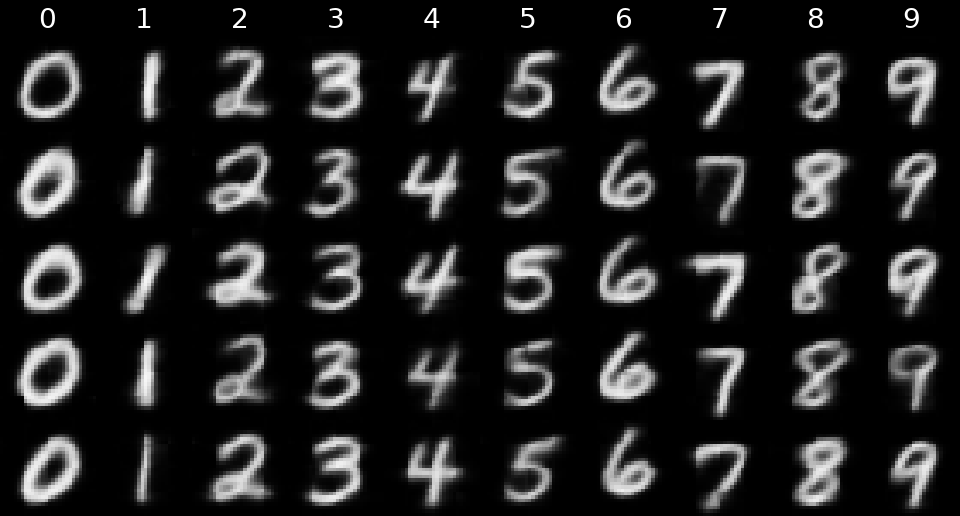}
        \caption{Unconditional text-and-image generation (step $18{,}500$).
                The digit word is generated before the image.}
        \label{fig:mnist-uncond}
    \end{subfigure}
    \caption{Qualitative multimodal generation on MNIST, with one forward
        pass per generated block.}
    \label{fig:mnist-pair}
\end{figure}

We demonstrate the same block-wise generation mechanism beyond text
in two multimodal settings (Fig.~\ref{fig:mnist-pair}). In
\emph{text-to-image generation}, an English digit word conditions an
image block predicted in one forward pass. In \emph{unconditional
text-and-image generation}, the model first generates a digit word and
then an image conditioned on that word, using one forward pass per
block. These MNIST examples provide a qualitative demonstration of
multimodal generation.

\FloatBarrier
\section{Conclusion}
\label{sec:conclusion}

We introduced \ours{}, a noise-conditioned masked diffusion model
trained from scratch with winner-take-all supervision and same-noise
self-distillation. The method learns to map different noise samples to
different continuations, predicting a complete block in one forward pass.
Small-scale experiments demonstrate single-block and block-wise text
generation, as well as text-to-image and unconditional text-and-image
generation.


\section{Limitations}
\label{sec:limitations}

The experiments here study TinyStories continuation and a qualitative
MNIST demonstration. Larger models, more complex and diverse datasets,
reliable long-form generation, and broader generalization are natural
directions for follow-up work.

\bibliographystyle{plainnat}
\bibliography{refs}

\clearpage
\appendix
\setcounter{table}{0}
\renewcommand{\thetable}{S\arabic{table}}
\setcounter{figure}{0}
\renewcommand{\thefigure}{S\arabic{figure}}
\section*{Supplementary Material}
Appendix~\ref{app:impl} provides architecture, training schedules, and the
DiMO adaptation. Appendix~\ref{app:judging} defines the judging rubric. Appendix~\ref{app:results} reports
additional experiments and examples.

\section{Additional experimental details}
\label{app:impl}

\subsection{Architecture and training details}
\label{app:training-details}

Table~\ref{tab:implementation-config} complements the core settings in
Table~\ref{tab:training-config}. The basic model uses pre-norm
layers, rotary position embeddings, and tied input/output token
embeddings. The teacher in the basic recipe is the current student.
The EMA row in Table~\ref{tab:main} instead uses decay $0.9999$,
initialized from the student and updated after each optimizer step.

\begin{table}[!ht]
\centering\small
\caption{Optimizer details for the basic \ours{} recipe.}
\label{tab:implementation-config}
\setlength{\tabcolsep}{4pt}
\fitwidth{\begin{tabular}{@{}ll@{}}
\toprule
AdamW $(\beta_1,\beta_2)$ & $(0.9,0.999)$ \\
Weight decay & $0.01$ \\
Warm-up updates $W$ & $1{,}333$ \\
Final learning rate & $0$ \\
\bottomrule
\end{tabular}}
\end{table}

\paragraph{Learning-rate and distillation schedules.}
Let $S=100{,}000$ denote total updates and $\eta_0=3\times10^{-4}$.
Training uses
$\eta_t=\eta_0\min(1,t/W)[1+\cos(\pi t/S)]/2$ at update $t$,
with $W=1{,}333$.
The distillation weight is
$w(t)=0.1+0.9\min(t/53{,}333,1)$.

\subsection{Baseline training}
\label{app:baseline-details}

\paragraph{Plain CE, IWAE, and WTA.}
These methods use the same learning rate, weight decay, and warm-up as the
basic recipe in Table~\ref{tab:training-config}, with uniform mask sampling
unless specified otherwise.
IWAE uses this same schedule and $k=4$.
Its objective is Eq.~\eqref{eq:iwae-w}.
IWAE with $\pi_j=1/k$ uses the same schedule with uniform weights.

\paragraph{DiMO.}
DiMO~\citep{zhu2025dimo} originally distills a pretrained teacher for
discrete image generation. We report two adaptations to eight-token text
continuation, using the same Transformer trunk and BPE vocabulary as the
basic text recipe. Student, frozen teacher, and fake model are initialized
from the $100$k-update Plain CE checkpoint.

The \emph{hybrid} adaptation retains the released MaskGIT branch's
input settings, $r_{\rm init}=0.6$ and $p_{\rm init}=0.1$.
For an eight-token block, rounding gives five MASK positions and three
uniformly sampled non-MASK tokens. All eight completion embeddings are
perturbed as $\sqrt{1-0.1^2}\,e+0.1\,\rms(E)\epsilon$, with independent
standard Gaussian $\epsilon$ at each position.
The \emph{all-MASK plus noise} adaptation is our input variant:
every completion position starts at $\emask+0.5\,\rms(E)\epsilon$.
Both retain the released MaskGIT configuration's student and fake learning
rates of $10^{-5}$ and freeze the token embedding and its tied output
head.\footnote{Released
\href{https://github.com/yuanzhi-zhu/DiMO/blob/main/configs/DIMO_MaskGit.yaml}{MaskGIT training configuration}.
These settings refer specifically to this branch of DiMO.}

Each update categorically samples a student continuation at temperature
$1$ and remasks positions using the arccos schedule. The teacher--fake
forward-KL gradient at remasked positions updates the student logits.
One fresh student sample trains the fake model with masked cross-entropy
and label smoothing $0.1$. No additional ground-truth CE or EMA is used
during distillation.
Our text training uses $100{,}000$ distillation updates, batch size
$4{,}096$, AdamW with $(\beta_1,\beta_2)=(0.9,0.999)$ and weight decay
$0.01$, gradient-norm clipping at $1$, and seed $0$.
Both learning rates follow the formula above with
$\eta_0=10^{-5}$, $S=100{,}000$, and $W=1{,}333$.
Parameters remain FP32, with BF16 forward autocast.
This training schedule is our text adaptation; the released MaskGIT
configuration uses Adam with zero weight decay, a constant learning rate
after $100$ warm-up updates, and EMA with decay $0.9999$.

\section{Evaluation protocols and metrics}
\label{app:judging}

\subsection{Frozen-rubric text-validity re-review}

\paragraph{Criterion.}
A continuation is judged together with its visible prefix.
Natural truncation is allowed.
Internal grammar errors, merged words, token loops, control tokens, and
punctuation degeneration fail.
A close-quote can pass when the visible text already reads as the end of
speech. An extra close-quote or a nesting the visible text already breaks
still fails.
Boundary items are concrete ambiguities, such as a name switch.
The reported \emph{mixed} policy uses each item's recorded decision.
The judge is a language model that had seen earlier rubric discussion, so
this is not an independent human review. Labels were locked before
aggregation.

\section{Additional experiments and examples}
\label{app:results}

\subsection{Parallel candidate selection: additional results}
\label{app:block-selection}

Table~\ref{tab:block-selection-kl} repeats the candidate-count experiment
of Section~\ref{sec:block} with the KL checkpoint. It measures the effect
of increasing parallel candidates and changing the selection score within
that checkpoint. Labels come from its separately frozen Codex review.
Both the main-text original sweep and this KL sweep use $10{,}000$
updates, with different training recipes and review rounds.
Noise-stream seeds are paired across settings, but candidate tokens can
differ slightly with batch size.

\begin{table}[!ht]
\centering\small
\caption{The same candidate-count and strategy sweep for the KL checkpoint at step $10{,}000$. The metric definitions and 16-trajectory panel match Table~\ref{tab:block-selection}; rates are computed from this checkpoint's separately frozen labels.}
\label{tab:block-selection-kl}
\setlength{\tabcolsep}{4pt}
\fitwidth{\begin{tabular}{rlrrrrrrrrr}
\toprule
& & & \multicolumn{8}{c}{First $M$ blocks all valid (\%)} \\
\cmidrule(lr){4-11}
$n$ & Score & \shortstack{Block validity\\(\%)} & $M=1$ & $2$ & $3$ & $4$ & $5$ & $6$ & $7$ & $8$ \\
\midrule
1 & Any & 84.38 & 93.75 & 93.75 & 87.50 & 56.25 & 43.75 & 31.25 & 31.25 & 31.25 \\
4 & $\bar\ell$ & 91.41 & 100.00 & 93.75 & 93.75 & 87.50 & 75.00 & 75.00 & 68.75 & 68.75 \\
4 & $\ell_{\min}$ & 90.62 & 100.00 & 87.50 & 81.25 & 75.00 & 68.75 & 68.75 & 68.75 & 56.25 \\
4 & $(\bar\ell+\ell_{\min})/2$ & 92.97 & 100.00 & 93.75 & 87.50 & 87.50 & 87.50 & 87.50 & 81.25 & 75.00 \\
8 & $\bar\ell$ & 93.75 & 100.00 & 100.00 & 100.00 & 100.00 & 75.00 & 75.00 & 75.00 & 68.75 \\
8 & $\ell_{\min}$ & 93.75 & 100.00 & 93.75 & 93.75 & 81.25 & 75.00 & 75.00 & 68.75 & 62.50 \\
8 & $(\bar\ell+\ell_{\min})/2$ & 93.75 & 100.00 & 93.75 & 87.50 & 87.50 & 75.00 & 75.00 & 75.00 & 75.00 \\
16 & $\bar\ell$ & 92.97 & 100.00 & 100.00 & 100.00 & 93.75 & 81.25 & 75.00 & 75.00 & 75.00 \\
16 & $\ell_{\min}$ & 89.84 & 100.00 & 100.00 & 100.00 & 87.50 & 87.50 & 75.00 & 75.00 & 62.50 \\
16 & $(\bar\ell+\ell_{\min})/2$ & 91.41 & 100.00 & 100.00 & 100.00 & 93.75 & 93.75 & 75.00 & 75.00 & 62.50 \\
\bottomrule
\end{tabular}}
\end{table}

\subsection{One-step DiMO adaptations for text}
\label{app:dimo}

Table~\ref{tab:dimo-results} evaluates the final $100{,}000$-update
students from Appendix~\ref{app:baseline-details} on the same frozen
$256$ held-out prefixes of random non-opening sentences used by the
single-block evaluation. Each prefix contains eight tokens and receives
$16$ initialization draws, with generation seed $0$.
We retain each adaptation's own input initialization and generate all
eight positions in one forward pass with BF16 weights and inputs.
For this text evaluation, outputs use token argmax; the released DiMO
MaskGIT sampler instead uses categorical sampling.\footnote{Released
\href{https://github.com/yuanzhi-zhu/DiMO/blob/main/sample_MaskGit.py}{MaskGIT one-step sampler}.}
Visible outputs stop at the first EOS or PAD.

The columns follow Section~\ref{sec:metrics}.
Validity, $V_{\ge 1}$, Uniqueness \& Validity, and their ratio use the
first $32\times16$ blocks; Uniqueness and gen-PPL use all
$256\times16$ blocks.
Uniqueness compares decoded strings exactly.
GPT-2-large uses FP32 weights and cross-entropy, with eager attention and
TF32 disabled. Prefix and continuation are jointly retokenized;
tokens overlapping the continuation contribute to pooled NLL, and
gen-PPL is $\exp(\text{total NLL}/\text{total scored tokens})$.
Validity labels come from a source-anonymized Codex review under the
mixed rubric in Appendix~\ref{app:judging}.
The same assistant organized and judged the review, having previously
seen the configurations and a few demonstration outputs; labels were
locked before unblinding.

\begin{table}[!ht]
\centering\small
\caption{DiMO text adaptations on random-sentence eight-token prefixes,
using the main-table metric definitions. Both rows use learning rate
$10^{-5}$ for student and fake models, with frozen embeddings and tied
output heads. All-MASK plus noise is our input variant; Hybrid retains the released
MaskGIT hybrid-input settings.}
\label{tab:dimo-results}
\setlength{\tabcolsep}{2pt}
\fitwidth{\begin{tabular}{@{}lrrrrrrr@{}}
\toprule
Input variant & NFE & \shortstack{Validity\\(\%) $\uparrow$} &
\shortstack{$V_{\ge 1}$\\(\%) $\uparrow$} &
\shortstack{Uniqueness\\(\%) $\uparrow$} &
\shortstack{U\&V\\(\%) $\uparrow$} &
\shortstack{(U\&V)/V\\(\%) $\uparrow$} &
gen-PPL $\downarrow$ \\
\midrule
All-MASK plus noise & 1 & 56.84 & 100.00 & 99.44 & 56.05 & 98.63 & 59.46 \\
Hybrid & 1 & 58.40 & 100.00 & 99.51 & 58.20 & 99.67 & 58.65 \\
\bottomrule
\end{tabular}}
\end{table}

\subsection{One-step flow adaptations for text}
\label{app:flow}

We adapt MeanFlow~\citep{geng2025meanflow} and rectified
flow~\citep{liu2023rectified} to the same eight-token-prefix,
eight-token-completion task. Each model is trained from scratch on
TinyStoriesV2-GPT4 with the byte-level BPE vocabulary of $8{,}192$.
Training windows begin at a uniformly sampled sentence boundary within
one story. The first eight tokens provide clean conditioning; the next
eight are the completion. The six models share the settings in
Table~\ref{tab:flow-config} and use training seed $0$.
The batch is $4{,}096$, compared with $512$ for the basic single-block
recipe in Table~\ref{tab:training-config}.

\begin{table}[!ht]
\centering\small
\caption{Shared settings of the six flow adaptations.}
\label{tab:flow-config}
\setlength{\tabcolsep}{4pt}
\fitwidth{\begin{tabular}{@{}ll@{}}
\toprule
Architecture & Block-causal Transformer with RoPE \\
Layers / hidden / FFN & $6/256/1024$ \\
Attention heads (query / key-value) & $8/8$ \\
Parameters, including flow heads & $8.66$M \\
Sequence / prefix / completion length & $16/8/8$ \\
Total updates / batch size & $100{,}000/4{,}096$ \\
AdamW $(\beta_1,\beta_2)$ / weight decay & $(0.9,0.999)/0.01$ \\
Peak learning rate / warm-up updates & $3\times10^{-4}/1{,}333$ \\
Learning-rate decay / gradient-norm clip & Cosine to zero after warm-up / $1.0$ \\
Flow / table CE / prediction CE weights & $1/1/1$ from the first update \\
Parameter / forward precision & FP32 / BF16 autocast \\
\bottomrule
\end{tabular}}
\end{table}

\paragraph{Shared embedding and prediction losses.}
Let $E\in\mathbb{R}^{8192\times256}$ be the learned token table,
initialized with independent $\mathcal{N}(0,256^{-1})$ entries and tied
to the output head. The Transformer retains the single-block attention
mask: prefix queries see only the prefix, while completion queries see
the prefix and the whole completion. A sinusoidal encoding of $(r,t)$
passes through a $512\to256\to256$ SiLU MLP and is added to the input
vectors. A zero-initialized $256\to256$ linear head predicts velocity.
For gold completion $y$, set $x=\operatorname{sg}(E[y])$ and
$\sigma_E=\sqrt{\operatorname{mean}(E^2)}$, and draw independent Gaussian
noise at each completion position, $\epsilon\sim
\mathcal{N}(0,\sigma_E^2 I)$. The path is
$z_t=(1-t)x+t\epsilon$, with data at $t=0$, noise at $t=1$, and
$v=\epsilon-x$.
The loss is the sum of flow regression, token-table self-classification
$\CE(E[p,y]E^\top,[p,y])$, and prediction CE
$\CE(\hat{x}E^\top,y)$. Table self-classification covers all sixteen
prefix and completion tokens. Flow and prediction inputs use detached
prefix embeddings, and flow endpoints are detached from the table.
The one-step prediction is $\hat{x}=\epsilon-u_\theta(\epsilon,0,1)$ for
MeanFlow and $\hat{x}=\epsilon-v_\theta(\epsilon,1,1)$ for rectified flow.
There is no CE warm-up or self-distillation teacher.

\paragraph{Flow objectives.}
Rectified flow samples $t=\operatorname{sigmoid}(a)$ with
$a\sim\mathcal{N}(-0.4,1)$, sets $r=t$, and minimizes the mean squared
velocity error divided by $\sigma_E^2$.
MeanFlow draws two such times, orders them as $r\leq t$, and sets $r=t$
for half the batch. Its detached target is
$v-(t-r)\,\mathrm{d}u_\theta/\mathrm{d}t$, where the directional
derivative follows $(z,r,t)$ in direction $(v,0,1)$.
If $q$ is the mean squared error against this target for one sample,
the adaptive regression loss is
$\operatorname{sg}[(q+10^{-3})^{-1}]q$, averaged over the batch.

\paragraph{The six settings.}
Each objective has three prediction-CE gradient routes
(Table~\ref{tab:flow-variants}). One updates both the velocity
network and the tied table. One detaches $\hat{x}$ and updates
only the table. One detaches the output-head weight and updates
only the velocity network. These stops affect prediction CE alone.
In every setting, flow regression updates only the velocity network:
prefix embeddings, completion endpoints, and the noise scale are detached
from $E$. The table is trained by table self-classification, and also by
prediction CE except in the network-only route, where the output head is detached.

\begin{table}[!ht]
\centering\small
\caption{Settings that differ across the six flow adaptations. All other settings are shared in
Table~\ref{tab:flow-config}.}
\label{tab:flow-variants}
\setlength{\tabcolsep}{5pt}
\fitwidth{\begin{tabular}{@{}ll@{}}
\toprule
Flow objective & Prediction CE updates \\
\midrule
MeanFlow & Network and table \\
MeanFlow & Table only \\
MeanFlow & Network only \\
Rectified flow & Network and table \\
Rectified flow & Table only \\
Rectified flow & Network only \\
\bottomrule
\end{tabular}}
\end{table}

\paragraph{Evaluation and results.}
We evaluate the final $100{,}000$-update checkpoints with one network
evaluation per eight-token block. Each of $256$ fixed held-out prefixes
contains the first eight tokens of a sampled non-opening sentence;
each prefix receives $16$ Gaussian draws. Generation seed is $0$, and
the six models use the same unit Gaussian draws scaled by their own
$\sigma_E$. All eight positions are decoded by token argmax.
The metric definitions follow Section~\ref{sec:metrics}: Validity,
$V_{\ge 1}$, Uniqueness \& Validity, and their ratio use the first
$32\times16$ blocks, while Uniqueness and gen-PPL use all
$256\times16$ blocks.
Uniqueness compares decoded strings exactly. GPT-2-large scores the
generated continuation after jointly tokenizing prefix and continuation;
gen-PPL is the exponential of total continuation NLL divided by the
number of scored tokens. The scorer uses BF16 weights and FP32
cross-entropy.
Validity labels come from a source-anonymized Codex review under the
mixed rubric in Appendix~\ref{app:judging}. The reviewer had prior
exposure to these experiments; the evaluation cards were shuffled and
the labels were locked before aggregation.

\begin{table}[!ht]
\centering\small
\caption{Final one-step flow results using the main-table metric
definitions. Rows match Table~\ref{tab:flow-variants}.
(U\&V)/V is undefined when there are no valid blocks.}
\label{tab:flow-results}
\setlength{\tabcolsep}{2pt}
\fitwidth{\begin{tabular}{@{}lrrrrrrr@{}}
\toprule
Adaptation & NFE & \shortstack{Validity\\(\%) $\uparrow$} &
\shortstack{$V_{\ge 1}$\\(\%) $\uparrow$} &
\shortstack{Uniqueness\\(\%) $\uparrow$} &
\shortstack{U\&V\\(\%) $\uparrow$} &
\shortstack{(U\&V)/V\\(\%) $\uparrow$} &
gen-PPL $\downarrow$ \\
\midrule
MeanFlow, network and table & 1 & 1.37 & 9.38 & 42.92 & 0.59 & 42.86 & 275.97 \\
MeanFlow, table only & 1 & 0.00 & 0.00 & 100.00 & 0.00 & --- & 2342.49 \\
MeanFlow, network only & 1 & 0.20 & 3.13 & 50.24 & 0.20 & 100.00 & 307.87 \\
Rectified flow, network and table & 1 & 1.76 & 18.75 & 42.55 & 1.37 & 77.78 & 248.48 \\
Rectified flow, table only & 1 & 0.00 & 0.00 & 96.85 & 0.00 & --- & 1657.57 \\
Rectified flow, network only & 1 & 1.37 & 6.25 & 51.37 & 0.59 & 42.86 & 296.29 \\
\bottomrule
\end{tabular}}
\end{table}

Under this recipe, one-step validity is at most $1.76\%$ ($9/512$ blocks),
for rectified flow with prediction CE on both the network and the table.
The table-only rows have no valid blocks in the judged panel,
so their ratio is undefined, while Uniqueness stays near $100\%$.
For MeanFlow with prediction CE on the network only, (U\&V)/V is $100\%$
because its single valid block is one string;
$V_{\ge 1}$ is $3.13\%$, one prefix out of $32$.

\subsection{Noise geometry}
\label{app:noisegeometry}

To test how inference noise affects the continuation, we hold the block-wise
partial-mask checkpoint at step $10{,}000$ and the text prefixes fixed, and
change only the noise (Table~\ref{tab:noise_radius}).
Each continuation is one eight-token block from one forward pass, with no retraining.
The noise is one shared vector of dimension $H=1024$, added at all eight mask
positions.
$r_0$ is the root-mean-square norm of the shared noise vector during training.
On this checkpoint the learned per-coordinate scale is $\sigma=0.07958984375$
and $r_0=\sigma\sqrt{H}=2.546875$.
A radius-$\alpha r_0$ input keeps a unit direction $u$ and sets
$\delta=\alpha r_0 u$, so $\|\delta\|_2=\alpha r_0$.
The radius sweep keeps each noise direction and sets the norm to a multiple
of $r_0$.
Automatic scores use $128\times 64$ continuations, $64$ draws for every one
of $128$ prefixes, and report Uniqueness for those draws.
This sweep does not re-judge Validity.
An angle-$\theta$ input has the same norm and angle $\theta$ to a reference
direction, with the tangent uniform on the orthogonal complement.
The angle sweep keeps the norm at $r_0$ and sets the angle to a reference
noise on the same prefix.
The $0^\circ$ and $180^\circ$ cases are a single direction.
The reference output is the block generated from that reference noise.
Within each reference group, pairwise token disagreement is the mean
fraction of token positions that differ between output pairs, and reference
change is the fraction of blocks that differ from the reference output in
any token.
Conf.\ is the mean model probability of the generated tokens.
$\log P$ is the sum of the log-probabilities of the eight generated tokens.

Under that probe, a larger angle from the reference noise increases the
fraction of outputs that differ from the reference output, while gen-PPL
stays stable (Table~\ref{tab:noise_radius}).
A larger norm produces more distinct outputs, but an excessive norm lowers
confidence and raises gen-PPL.
Tables~\ref{tab:noise_norm_text} and~\ref{tab:noise_text} separate those two
effects on one prefix: the radius table keeps one direction, and the angle
table keeps the norm at $r_0$.
Both the direction and the magnitude of the noise affect the continuation.

\begin{table}[!ht]
\centering
\small
\caption{Inference-noise norm and direction on the block-wise partial-mask
checkpoint at step $10{,}000$. Each cell is one eight-token block.
Validity is not re-judged.}
\label{tab:noise_radius}
\label{tab:noise_angle}
\setlength{\tabcolsep}{2pt}
\fitwidth{\begin{tabular}{@{}lrrr|rrrrrr@{}}
\toprule
\multicolumn{4}{c|}{\emph{Fixed radius}} & \multicolumn{6}{c}{\emph{Fixed angle, norm $r_0$}} \\
\midrule
& \multicolumn{3}{c|}{Automatic: $128\times64$} & \multicolumn{6}{c}{Within each reference cone} \\
\cmidrule(lr){2-4}\cmidrule(lr){5-10}
Radius & \shortstack{Uniqueness\\(\%)} & \shortstack{Conf.\\(\%)} & \shortstack{gen-PPL\\$\downarrow$} & Angle & \shortstack{Uniqueness\\(\%)} & \shortstack{Pairwise token\\disagreement (\%)} & \shortstack{Reference\\change (\%)} & \shortstack{Conf.\\(\%)} & \shortstack{gen-PPL\\$\downarrow$} \\
\midrule
Original Gaussian & 37.90 & 64.6 & 20.80 & $0^\circ$ & 1.56 & 0.0 & 0.0 & 64.7 & 20.85 \\
$0\times r_0$ & 1.56 & 67.1 & 18.49 & $5^\circ$ & 7.28 & 14.7 & 31.1 & 64.8 & 20.44 \\
$0.25\times r_0$ & 22.99 & 66.3 & 19.58 & $15^\circ$ & 15.58 & 31.1 & 56.3 & 64.8 & 20.32 \\
$0.5\times r_0$ & 31.51 & 65.4 & 20.31 & $30^\circ$ & 24.77 & 42.6 & 72.4 & 64.8 & 20.31 \\
$0.75\times r_0$ & 35.55 & 65.0 & 20.72 & $45^\circ$ & 30.81 & 48.3 & 80.2 & 64.7 & 20.42 \\
$1\times r_0$ & 38.20 & 64.6 & 20.93 & $60^\circ$ & 34.79 & 51.2 & 84.6 & 64.6 & 20.53 \\
$1.25\times r_0$ & 39.26 & 64.2 & 21.03 & $90^\circ$ & 37.60 & 53.3 & 89.3 & 64.6 & 20.81 \\
$1.5\times r_0$ & 42.33 & 63.1 & 21.50 & $120^\circ$ & 34.42 & 50.9 & 90.9 & 64.5 & 20.59 \\
$2\times r_0$ & 52.95 & 57.7 & 28.01 & $150^\circ$ & 25.03 & 42.3 & 92.2 & 64.3 & 20.78 \\
$3\times r_0$ & 75.16 & 44.9 & 62.15 & $180^\circ$ & 1.56 & 0.0 & 92.0 & 64.5 & 20.50 \\
\bottomrule
\end{tabular}}
\end{table}

\begin{table}[!ht]
\centering
\small
\caption{Unedited continuations of \textit{It looked yummy, so he picked it} along one direction.
The $1\times r_0$ row is the $0^\circ$ reference in Table~\ref{tab:noise_text}.}
\label{tab:noise_norm_text}
\begin{tabular}{lp{7.2cm}rr}
\toprule
Radius & Generated continuation & Conf. (\%) & $\log P$ \\
\midrule
$0\times r_0$ & \texttt{up and said, "Thank you,} & 74.2 & $-2.480$ \\
$0.5\times r_0$ & \texttt{up and said, "Thank you,} & 68.9 & $-3.116$ \\
$1\times r_0$ & \texttt{up and said, "Thank you,} & 70.9 & $-2.871$ \\
$1.5\times r_0$ & \texttt{up and took it to bite.\textbackslash n} & 52.6 & $-5.966$ \\
$2\times r_0$ & \texttt{up his to.....} & 18.6 & $-14.693$ \\
$3\times r_0$ & \texttt{...,,,,.} & 27.0 & $-12.032$ \\
\bottomrule
\end{tabular}
\end{table}

\begin{table}[!ht]
\centering
\small
\caption{Unedited continuations of \textit{It looked yummy, so he picked it} at norm $r_0$.
Panel A follows one tangent; panel B uses other directions at $30^\circ$.}
\label{tab:noise_text}
\begin{tabular}{llp{7.0cm}rr}
\toprule
Angle & Tangent & Generated continuation & Conf. (\%) & $\log P$ \\
\midrule
\multicolumn{5}{l}{\emph{A. One paired rotation trajectory}} \\
$0^\circ$ & --- & \texttt{up and said, "Thank you,} & 70.9 & $-2.871$ \\
$30^\circ$ & $v_0$ & \texttt{up and put it in it. It} & 59.4 & $-5.134$ \\
$90^\circ$ & $v_0$ & \texttt{up and said, "I want to} & 75.8 & $-2.374$ \\
$180^\circ$ & --- & \texttt{up and put it in a big,} & 71.1 & $-3.424$ \\
\midrule
\multicolumn{5}{l}{\emph{B. Multiple directions with the same angle to the same reference}} \\
$30^\circ$ & $v_1$ & \texttt{up and said, "Thank you,} & 74.4 & $-2.457$ \\
$30^\circ$ & $v_2$ & \texttt{up and started to eat it.\textbackslash n} & 65.6 & $-3.606$ \\
$30^\circ$ & $v_4$ & \texttt{up and took it in bite.\textbackslash n} & 57.8 & $-5.343$ \\
$30^\circ$ & $v_6$ & \texttt{up and said, "Thank you,} & 77.8 & $-2.075$ \\
\bottomrule
\end{tabular}
\end{table}

\end{document}